\documentclass[letterpaper, 10 pt]{ieeeconf}  

\IEEEoverridecommandlockouts                              

\usepackage{graphics} 
\usepackage{epsfig} 
\usepackage{mathptmx} 
\usepackage{times} 
\usepackage{amsmath} 
\usepackage{amssymb}  
\usepackage{verbatim} 
\usepackage{xspace}
\usepackage[subtle]{savetrees}

\usepackage{cleveref} 

\newcommand{\algoName}[0]{See-PP\xspace}

\title{\LARGE \bf
See, Plan, Predict: Language-guided Cognitive Planning with Video Prediction
}

\author{Maria Attarian$^{1,2,3*}$\thanks{*equal contribution}, Advaya Gupta$^{1,2*}$, Ziyi Zhou$^{1,2}$, Wei Yu$^{1,2}$, Igor Gilitschenski$^{1}$, Animesh Garg$^{1,2}$
\thanks{$^{1}$University of Toronto, $^{2}$Vector Institute, $^{3}$Google}%
}

\begin{document}

\maketitle

\thispagestyle{empty}
\pagestyle{empty}

\begin{abstract}
Cognitive planning is the structural decomposition of complex tasks into a sequence of future behaviors. In the computational setting, performing cognitive planning entails grounding plans and concepts in one or more modalities in order to leverage them for low level control. Since real-world tasks are often described in natural language, we devise a cognitive planning algorithm via language-guided video prediction. Current video prediction models do not support conditioning on natural language instructions. Therefore, we propose a new video prediction architecture which leverages the power of pre-trained transformers. The network is endowed with the ability to ground concepts based on natural language input with generalization to unseen objects. We demonstrate the effectiveness of this approach on a new simulation dataset, where each task is defined by a high-level action described in natural language. Our experiments compare our method against one video generation baseline without planning or action grounding and showcase significant improvements. Our ablation studies highlight an improved generalization to unseen objects that natural language embeddings offer to concept grounding ability, as well as the importance of planning towards visual "imagination" of a task.
\end{abstract}

\section{INTRODUCTION}

\begin{figure}[thpb]
  \centering
  \framebox{\parbox{3in}{\includegraphics[width=\linewidth]{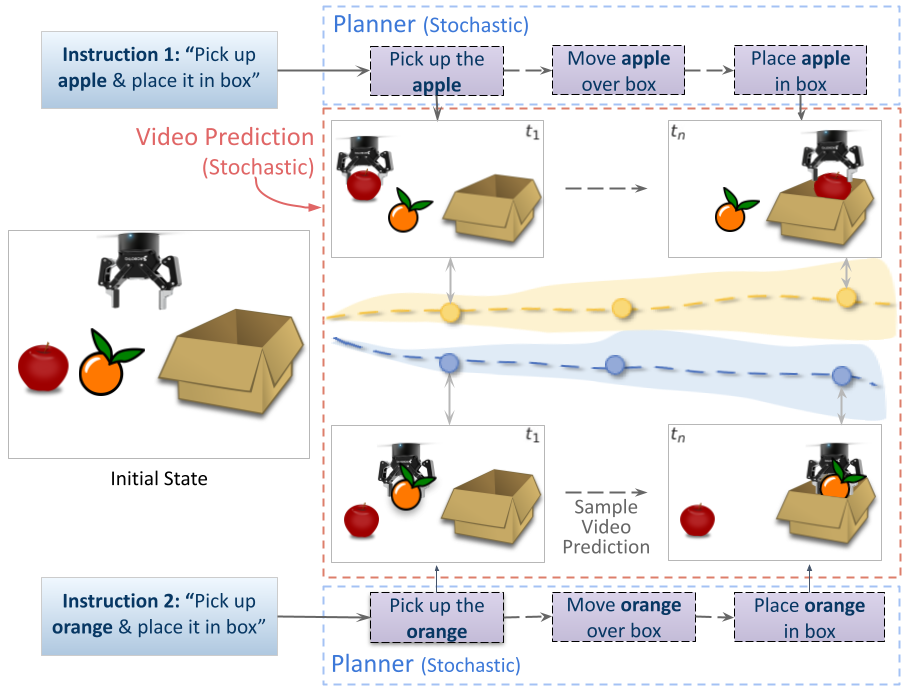}}}
  \caption{Intuition behind cognitive planning as performed by humans. A high level task is broken down into steps with each performed in sequence. The subgoals of the various steps are met in the process leading up to the final goal.}
  \label{fig:main-fig}
\end{figure}

Cognitive planning is one of the core abilities that allows humans to carry out complex tasks through formulation, evaluation and selection of a sequence of actions and expected percepts to achieve a desired goal~\cite{cognitiveplanning1997}. The ability to look ahead and to conditionally predict the expected perceptual input is necessary for goal-conditioned planning. However, at times the intermediate steps involved may not directly relate to achieving this goal. Consider the scenario illustrated in Fig.~\ref{fig:main-fig}: We have two fruits and we would like to place one in the box. This may elicit the thought to “pick up the apple”, “move it over the box”, and “place the apple inside”. Such a plan may also trigger visual and other sensory associations when planning the corresponding actions with only an approximate world-model. As put, cognitive planning can be thought of as a combination of two tasks: (i) high-level planning with abstract actions, and (ii) concept grounding of the planned sequence of actions. This is subsequently followed by physical execution of the abstract action sequence with feedback-control using the grounding as reference. We thus approach the problem of task completion as having three phases: (a) high-level planning, (b) cognitive grounding, and (c) closed-loop control. In this work, we address the first two phases of this pipeline and assess our ability to perform video generation via natural language instruction and conceptual reasoning about a scene. Applying our results towards low level robotic control is left as future work. 

Cognitive grounding has multiple forms~\cite{cognitivegrounding2013}. For our example this takes the form of grounding concepts in the visual space. What does “apple” represent? What does “move it over the box” entail? We are grounding these concepts in vision when we imagine what they look like. Such “imagination” of future observations of subgoals
could be leveraged by various Imitation Learning approaches like observation-only behavior cloning~\cite{torabi2018behavioral}, LbW-kP~\cite{xiong2021learning} and Transporter Nets~\cite{zeng2020transporter} to perform these tasks. This motivates us to implement a computational analog of cognitive planning, which, in our three phase model, performs the first two phases: planning and grounding.

\begin{figure*}[thpb]
  \centering
    \framebox{\includegraphics[width=0.9\linewidth]{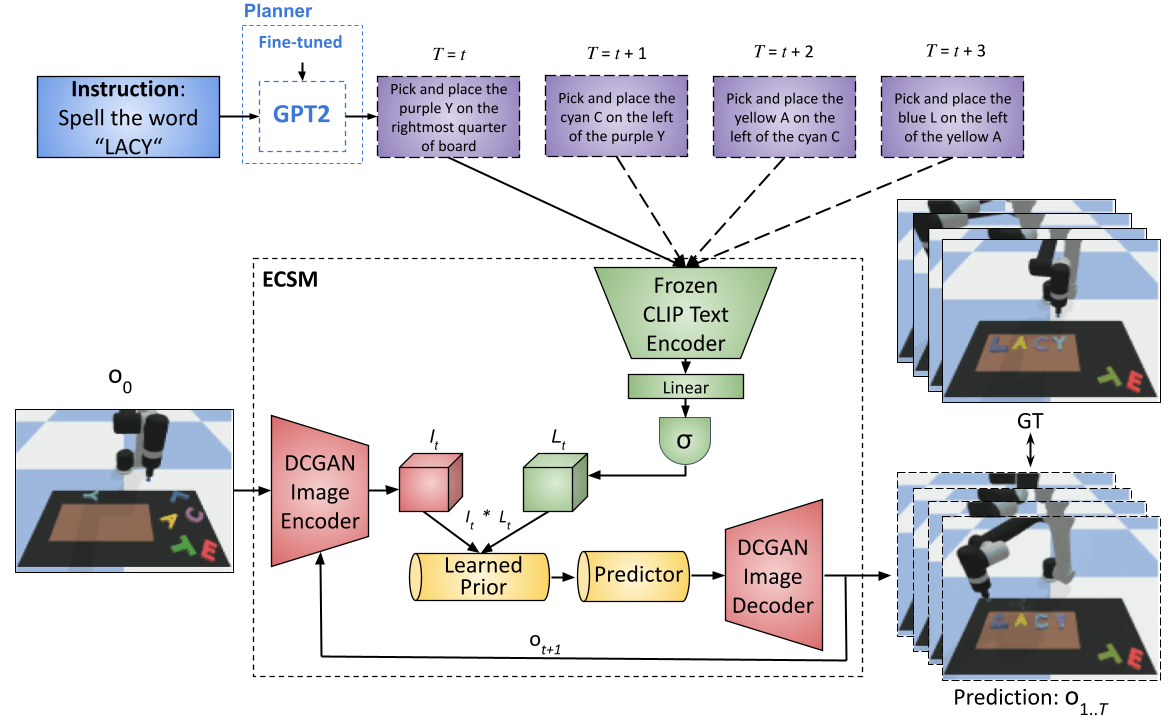}}
    \caption{Architecture diagram of \algoName. The high level description of the task is passed through the planner which suggests an ordered list of sequential goal instructions. The first of those is converted to a language embedding and combined with the image embedding of the initial observation, generated by the encoder. The predicted observation outputted by the predictor and decoder is used as the new initial frame to be combined with the next language instruction. This process is repeated until the end of the step-by-step plan.}
    \label{fig:diagram}
\end{figure*}

Concretely, our work proposes a novel architecture for performing cognitive planning based on an initial visual observation and a natural language task description. This involves producing a high-level plan of actions by generating natural language commands, which are used for visual grounding through language-conditioned video prediction. An architectural diagram of our method is illustrated in \cref{fig:diagram}.

\noindent \textbf{Summary of Contributions.}
\begin{enumerate}
\item A novel architecture that combines the planning power of pre-trained transformer models with visual concept grounding to tackle computational cognitive planning as a video prediction problem.

\item A simulation dataset of a robotic arm performing spelling of various 4-letter words on a board.

\item Evaluations that demonstrate the value of such a framework towards better semantic generalization in video prediction. 
\end{enumerate}

\section{RELATED WORK}

\noindent\textbf{Video prediction and generation.} Video prediction and generation has been extensively studied in the computer vision community. ConvLSTMs have been used for learning representations of video sequences on patches of image pixels as well as high-level representations~\cite{srivastava2015unsupervised}, while they've also been proven effective for video prediction on settings where objects are controlled or influenced by actions~\cite{oh2015action}. Hierarchical and non-hierarchical VAEs~\cite{denton2018stochastic, castrejon2019improved, wu2021greedy} are another interesting direction of approaches that have yielded state-of-the-art results in video prediction. ~\cite{finn2016unsupervised} conduct pixel transformations from previous frames, explicitly modeling motion where~\cite{vondrick2016generating} attempts to segregate foreground from background with the use of generative adversarial networks. Stochastic video prediction like~\cite{denton2018stochastic, lee2018stochastic, babaeizadeh2017stochastic} has yielded advancements in the field, however these approaches neither support language instructions nor provide a way to control predictions based on text directives. 
Such level of control is not provided by existing stochastic video prediction models which instead, predict different possible futures with no determinism based on a task definition. In comparison, approaches such as~\cite{yu2021modular} support tokenized instructions for relatively simpler task settings while also assuming access to such ground truth task segmentation. 
Finally, other approaches draw inspiration from language models~\cite{ranzato2014video, sun2019videobert, yan2021videogpt}. Similar to some prior work, our approach employs VAEs but operates on visual and language representations combined, attempting to do feature fusion from the two modalities. \newline
\noindent\textbf{Concept and action grounding.} Concept learning draws inspiration from Cognitive Science and attempts to study how humans conceptualize the world~\cite{cognitivereasoning2019}. Many approaches study concept and action grounding by operating directly in the image space or implicitly on some latent space~\cite{locatello2020object, zhou2019grounded}. Even though such methods have shown promising results, we follow the vein of approaches that explore the combination of language and vision representations for concept and action learning such as~\cite{gao2016acd, he2019read, migimatsu2021grounding}. Such approaches received even more attention after the emergence of the CLIP architecture for learning visual concepts from natural language supervision~\cite{radford2021learning}. Our intuition around concept and action grounding is related the closest to~\cite{shao2020concept2robot}. In this work, a shared embedding is generated by the concatenation of a reduced dimensionality language and vision embedding derived from large pre-trained state-of-the-art models. This shared embedding is subsequently employed to conceptually represent the scene. \newline
\noindent\textbf{Task planning.} Task and procedure planning is defined as the problem of predicting a sequence of actions that will accomplish a goal, when starting from some initial state. Several recent works attempt to perform planning from pixel observations. InfoGAN~\cite{chen2016infogan} seeks to learn causality in the data structure and extracts latent features that can then be used as state representations, whereas PlaNet~\cite{hafner2019learning} performs action planning in latent space via a model-based agent that learns its environment dynamics. UPN~\cite{srinivas2018universal} directly predicts an action plan through gradient descent trajectory optimization in the latent space. Another stream of approaches such as~\cite{chang2020procedure} studies planning in real-world instructional videos. Finally, several methods combine vision and language towards grounded planning~\cite{sun2021plate, blukis2021persistent, zhang2021hierarchical}. In contrast, our work segments the instructional and visual planning into two modularized subproblems similar to~\cite{bi2021procedure}. However, our approach differs from~\cite{bi2021procedure} in that our generation model learns to “imagine” its future states rather than choose them from a defined set of possible states, where for instruction planning we are focusing on purely language based planning as per~\cite{jansen2020visually}.

\section{OUR METHOD: SEE, PLAN, PREDICT}

The ability to generate plans for the future is a basic requirement for carrying out many complex behaviors~\cite{cognitiveplanning1997}. This involves high-level planning which entails structurally decomposing a complex task into simpler intermediate steps, which according to ~\cite{cognitivegrounding2013} may not necessarily lead towards the end goal. 
In fact, Owen~\cite{cognitiveplanning1997} argues that cognitive planning is necessary when tasks have sequential structures and when “novel courses of action” must be generated and performed. Taking inspiration from this argument, we explore the necessity of high-level structural decomposition of complex tasks towards "hallucination" of how such tasks would be executed.

Combining this with the idea that abstract concepts are grounded in “a variety of bodily, affective, perceptual, and motor processes”~\cite{cognitivegrounding2013}, we explore the idea of grounding abstract natural language concepts in the visual domain. 
We anticipate that visual grounding can be directly leveraged for low level robotic control through its application on various methods~\cite{zeng2020transporter, torabi2018behavioral}.

\begin{figure*}[thpb]
  \centering
        \framebox{\includegraphics[width=\linewidth]{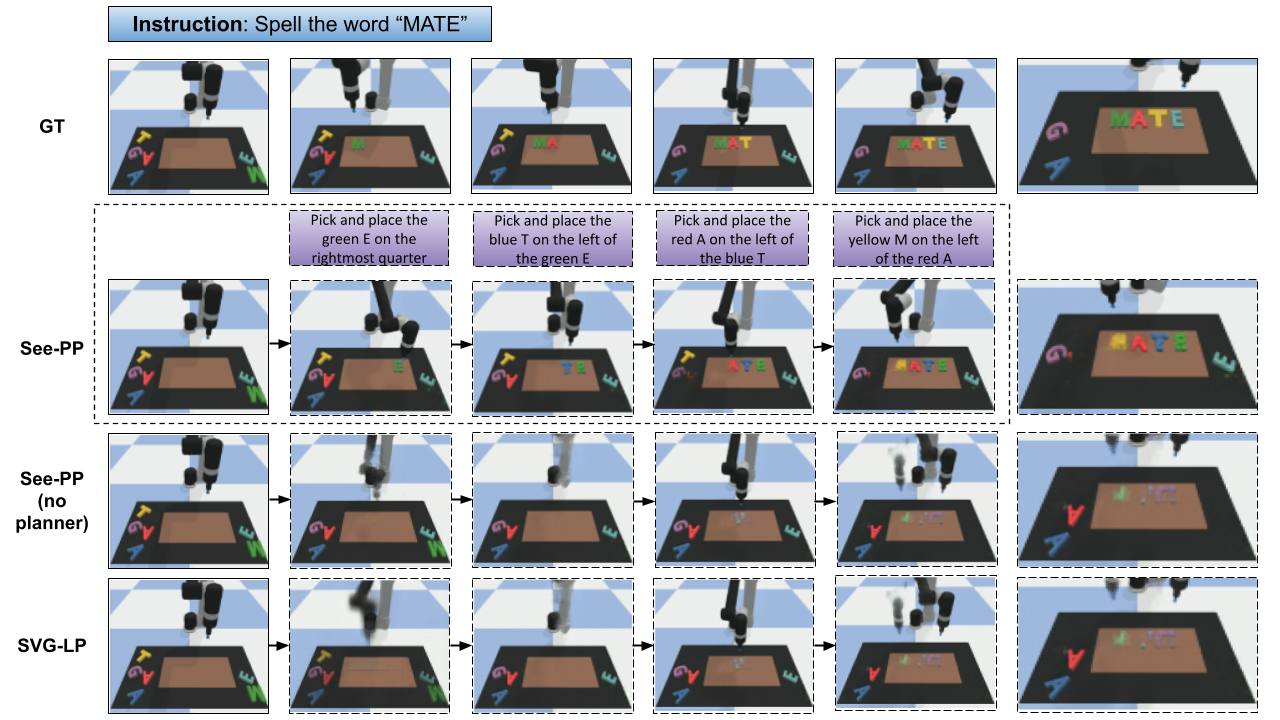}}
    \caption{Results for task "Spell the word MATE". \algoName is able to provide reasonable actions despite the blurriness of the objects. SVG-LP~\cite{denton2018stochastic} as well as \algoName in the absence of a planner are unable to even start the task.}
    \label{fig:svg-See-PP-ablations}
\end{figure*}


\subsection{Problem Definition}

We are interested in a similar problem to semantic action-conditioned video prediction~\cite{yu2021modular}. However, we do not assume access to a sequence of actions $a_{1:T}$ but rather explore approaches that only have access to the high level description of a task at inference time. 

Thus, our problem definition is as follows: Given a high level task description and an initial image observation $o_0 \in \mathbb{R}^{H\times W \times 3}$, can there be a model that predicts future observations $o_{1:T}$ that demonstrate a successful completion of a sequence of intermediate goals and ultimately the completion of said task? 
We divide this process into two subproblems: a) high level planning i.e. constructing a valid plan for the given high level task and b) video generation i.e. combining the instruction plan with the initial observation towards predicting a demonstration of successful completion of the task. In this context, a valid plan is defined as an ordered list of instructions that, if followed, can lead to the desired goal from the given initial state. While similar decompositions is common in the literature, many approaches focus more on planning from visual observations~\cite{bi2021procedure, chang2020procedure} or a visual-language combined aspect~\cite{sun2021plate, blukis2021persistent, zhang2021hierarchical}. In contrast, we focus on leveraging natural language only with the intuition being that humans can deconstruct a high level task to a valid instructional plan and execute it, without the use of visual demonstrations for each intermediate action outcome.

We present a two-fold approach which decomposes a language-based task instruction into short-horizon intermediate subgoals, and use this decomposition in order to ground the task concepts through video prediction. Moving forward, to assist the reader we shall refer to our method as \algoName in lieu of "See-Plan-Predict".

\subsection{Planning from High Level Descriptions}

Recent breakthroughs in the field of Natural Language Processing with the surge of Transformer architectures~\cite{vaswani2017attention} have led to a variety of Large Language Models (LLM) such as GPT2~\cite{radford2019language}, BERT~\cite{devlin2018bert}, RoBERTa~\cite{liu2019roberta} that provide powerful text representations and have achieved state-of-the-art performance in many language tasks. 
Building on LLMs, we frame our subproblem of deconstructing a high-level goal description to a list of subgoals as a strictly language-only task, similar to~\cite{jansen2020visually}. Using the Alfred benchmark~\cite{shridhar2020alfred} as a platform, this work demonstrated that the best performing LLM can achieve a 26\% success rate upon producing a plan for unseen tasks. For comparison, this is higher or comparable to the success rates on unseen tasks reported on the Alfred leaderboard, many of which address planning as a joint vision-language problem~\cite{blukis2021persistent, zhang2021hierarchical}. 

For our approach, we chose OpenAI GPT2 which was created for the purpose of text generation, and was demonstrated in~\cite{jansen2020visually}  as the best performing LLM for such a planning task. Similar to them, we fine-tuned pre-trained GPT2-medium on sequences of language high-level task descriptions followed by their ground truth step-by-step action commands, separated by delimiters e.g.  

\noindent {\small \texttt{$\langle$HighLevelTaskDescription$\rangle$[SEP]$\langle$ActionVerb$\rangle$ [ARG1] $\langle$ColorLetter$\rangle$[ARG2]$\langle$Position$\rangle$ $\langle$ColorLetter$\rangle$[CSEP]...[CSEP][EOS]}}. 

At evaluation time, the fine-tuned model is given the language directive of the high level task description followed by the separator delimiter [SEP] and produces a list of language action commands. More details on evaluation of the full method to follow.

\subsection{Extended Modular Action Concept Network}

Given a video frame, we aim to learn a mapping:
\begin{equation}
  \hat{o_t} = D(P(C^{\prime}(E(o_{t-1} | L_{t}) | h_{t-1})))
  \label{eq:exMAC}
\end{equation}

\noindent where $o_{t-1}$ is the video frame at time $t-1$, $L_t$ is the action instruction label corresponding to frame $t$, $h_{t-1}$ is the hidden state at time $t-1$, $E$ is the encoder model, $D$ is its corresponding decoder, $P$ is the predictor and $C^{\prime}$ is our proposed extension of the original CSM~\cite{yu2021modular}. 
The VAE setup in \algoName also includes a learned prior $p(z)$ and posterior $q(z)$ as part of the learning process, in the spirit of generalization in order to better handle perturbations of actions in our video generation through latent distribution approximation.  

\noindent \textbf{Encoder-Decoder:} With Generative Adversarial Networks (GAN) being at the heart of producing visual data via a generator, realistic enough to fool a discriminator, was philosophically and theoretically a straightforward candidate. Therefore, a DCGAN~\cite{radford2015unsupervised} inspired architecture was employed as the encoder-decoder of the VAE setup. 

\noindent \textbf{Extended Concept Slot Module (ECSM):} The goal of the CSM as described in~\cite{yu2021modular} is to ultimately have one slot per concept in the space of action labels. For clarity, it is reminded that a concept can be referring to any conceptual entity of a sentence, e.g. an action verb “pick” or an object phrase “green cup”. The slots corresponding to all concepts that comprise an action label, are then activated to represent said action. The action label decomposition into conceptual entities aims to create a representation of the language input. Each part of speech (PoS) is then assigned a unique one-hot encoding in a pre-defined dictionary of concepts and each label ultimately results in a one-hot encoding with the bit values of the representation corresponding to certain conceptual entities. These action label encodings are further combined with the feature map extracted from the input frame and used to activate the appropriate concept slots for each conceptual entity. 

The core idea around ECSM is founded on the observation that the number of concepts handled being finite and predefined through the use of a concept dictionary, which is a limitation of the above approach. Assuming a priori knowledge of all concepts that could possibly be encountered is not always feasible or realistic and introducing new concepts would require possibly re-parsing the entirety of one's data and retraining. In contrast, we seek to find an approach that eliminates this limitation and generalizes to some degree to unseen concepts. To this effect, and taking inspiration from recent advancements on learning relationships between visual input and natural language~\cite{radford2021learning}, we propose substituting action label decomposition based on predefined dictionaries with CLIP embeddings of action labels. With this, we aim to take advantage of the proximity in the embedding space that conceptually close CLIP text and image encodings already have been trained for. Alternatively, CLIP may be allowed to fine tune during training of ECSM, however for this work frozen pre-trainedCLIP embeddings were adequate to represent the conceptual space of our task.

In addition, given that the original CLIP embedding is too large for our application we reduced dimensionality by adding a learned linear projection layer as part of the gating functions. This layer trained in parallel with the overall video prediction algorithm. A sigmoid nonlinearity was further added after the linear projection layer as part of the language embedding. The intuition is that ECSM can be viewed as a continuous version of the CSM as described in~\cite{yu2021modular}. Continuous values for language embeddings can be obtained simply from the chosen pre-trained text encoder. However, an alternate definition can be obtained by transforming these continuous values into a probability range $[0, 1]$. The intuition for this is activation of the appropriate concept slots with some level of probability as opposed to binary activation. 
ECSM could therefore, be represented as: 
\begin{equation}
  {\bf w}^i = \Psi^i({\bf f}), \;\;\;    {\bf c}^j = \Phi^j (\mathrm{Concat}\{{\bf w}^j | \forall i \  {\bf emb} ^ i(j)\})
  \label{eq:exMAC-detail}
\end{equation}

\noindent where ${\bf w}$ and ${\bf c}$ are conceptual representations and ${\bf emb}$ is the embedding of the $j$th constituent. This method aims to leverage CLIP’s existing conceptual mapping of language to image space as an approach to jumpstart concept learning and concept grounding instead of training it from scratch as with predefined dictionaries. Our intuition lies within “fine-tuning”: is it possible to fine-tune concept learning similarly to LLMs or pre-trained vision model fine-tuning? 

\noindent \textbf{Learned prior, Predictor:} For these components, we did not modify the original implementation. 

\noindent \textbf{Training and Inference:} In the spirit of our modularized approach, training was held in two stages. First, we fine-tuned GPT2 for our spelling task. Subsequently, we trained the VAE setup of ECSM with a standard VAE loss comprised by a reconstruction loss component and a KL-divergence component:
\begin{equation}
  \mathcal{L}_{total} = \mathcal{L}_{reconstruct} + \mathcal{L}_{D_{KL}}
\end{equation}

Considering our dataset contains redundancy in regards to color, position and ordering of placement for a subset of words, our trained planner is multi-modal, i.e. it is expected to predict different plans for a given word. In addition, the inherent stochasticity of our dataset aims to further assist with variance handling and generalization capabilities. More details about our dataset follow. 

\begin{figure*}[thpb]
  \centering
     \framebox{\includegraphics[width=\linewidth]{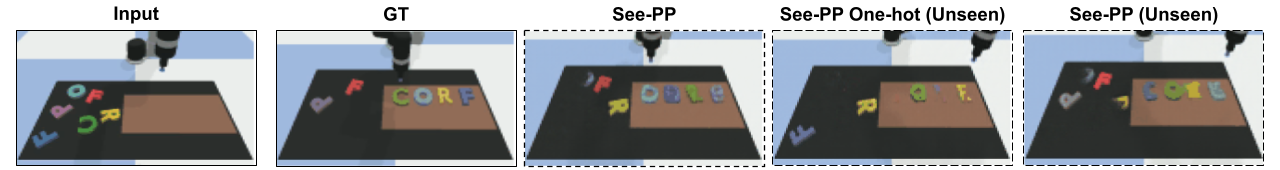}}
    \caption{\algoName on a test set containing unseen concepts versus a random split where all concepts in the test set are seen. We also compare to CSM's~\cite{yu2021modular} performance with planner (\algoName One-hot) on a test set with unseen concepts. The input instruction for the task shown is "Spell the word CORF." Our language planner generated the following series of instructions: "Pick and place the purple F on the rightmost quarter," "Pick and place the green R on the left of the purple F
    ," "Pick and place the blue O on the left of the green R
    ," "Pick and place the yellow C on the left of the blue O
    ." Leveraging the power of pre-trained LLMs for language representation enhances adaptation to new concepts. }
    \label{fig:seen-unseen}
\end{figure*}

\section{DATASET}

Our dataset design aimed to satisfy a number of properties that would allow us to examine the efficacy of our framework. A good dataset candidate for our purposes was deemed to have the following properties: 


 \begin{enumerate}
\item The dataset should be multimodal containing both visual and language information.
\item High level task descriptions that cannot be trivially broken down to goal instructions steps i.e. if “and” separated, the steps do not reconstruct the high level task description
\item Each action should be detailed enough that does not become trivial for a generation model to reconstruct. 
\item Enough variety of concepts should be present. 
\end{enumerate}

In addition to the desired properties taken into account for dataset design, the implicit exploration of relationships between visual concepts and natural language led us to a choice of domain where the visual observations are directly related to natural language. Furthermore, through our choice of CLIP as the text encoder for natural language representations, we were able to examine whether CLIP pre-trained language embeddings are able to map language to visual observations even when the latter are the language itself. In other words, whether CLIP pre-trained language representations have any structural or inherent association with visual presentation of language.

We modified the ravens environment~\cite{zeng2020transporter} to introduce the task of spelling four letter words. The scene includes 6 color-letter combinations chosen randomly from a set of 156 (26 letters and 6 possible colors), a workspace to spell the word in, and a robot arm. The task in each video is for the robot arm to spell the given four letter word using the letters provided in the scene. The objects are all visually unique, and are initialized with random positions and orientations in the scene. The colors of each letter are also randomly assigned. The dataset contains spelling demonstrations for 3368 distinct words. Demonstrations were collected via a scripted policy with randomized ordering of actions. For the purposes of our evaluation, we have created two types of training/test splits. The description and distribution of words for the two splits is as follows: \\ 
{\bf Train/test (random) split:} The full dataset is shuffled and videos are randomly divided into a training and test set. The train and test sets contain demonstrations for spelling 3178 distinct words over 4688 videos, and 442 distinct words over 466 videos respectively. \\
{\bf Seen/unseen split:} Training data is restricted to spelling words with only 22 letters, and the unseen 4 letters are only used for spelling words in the test set. The seen (train) and unseen (test) sets contain demonstrations for 2987 distinct words over 4655 videos, and 381 distinct words over 499 videos respectively. \\
Each video is accompanied by two types of labeling schemes: (i) a high-level action natural language label corresponding to the entire demonstration, (ii) low-level actions corresponding to each frame of the video. The high-level actions correspond to ``\texttt{Spell the word} $W_1$”. For the low-level actions, we have two types of experiments: one in which we use natural language action labels, and one in which the actions are encoded in the same format as ~\cite{yu2021modular}. The low-level actions correspond to actions like “Pick and place $O_1$ on the $P_1$ of the board” and “Pick and place $O_2$ to the $P_2$ of $O_1$”.

\section{EVALUATION}

To evaluate our work, we were interested the following questions: 

\noindent {\bf Q1:} How does our method compare to video generation without cognitive planning? 

\noindent {\bf Q2:} How important is the high level task planner for the overall learning task? How important is the extensible language representation for the overall learning task? 

\noindent {\bf Q3:} How does our method generalize? 

\subsection{Metrics}

We quantify the performance of our model by using MSE, PSNR, SSIM~\cite{wang2004image} and LPIPS~\cite{zhang2018unreasonable} metrics between predicted videos and ground truth. MSE and PSNR metrics quantify the strength of error signals, but assume pixel-wise independence. We therefore include SSIM to quantify structural similarity, and LPIPS~\cite{zhang2018unreasonable} as a perceptual similarity judgment. In addition, since our focus is in evaluating video prediction through the lens of performing a downstream task rather than the quality of the video itself, we calculated quality scores according to the method proposed in \cite{somraj2020pvqa}.

As quantitative metrics fail to effectively reflect whether actions are completed~\cite{zhang2018unreasonable, somraj2020pvqa}, we developed a customized Optical Character Recognition (OCR) system which is comprised by two classifiers that identify the character and its color respectively. We thus evaluated our generated videos based on the predictions of the OCR system after every action step for the models that incorporate planning and on the word level for all others. For clarity, it is noted that model predictions are purely video generation. The simulator used to produce ground truth data is not involved, therefore we cannot rely on known object positions and IDs. This justifies the use of an OCR system. We include qualitative examples of how our method performs in order to discuss visual observations on our method’s behavior. 

\begin{table}[h]
  \caption{Quantitative analysis of ablation studies for keyframe prediction.}
\small
  \centering
  \resizebox{\linewidth}{!}{%
  \begin{tabular}{|c| |c| |c| |c| |c| |c|}
    \hline
    Architecture & MSE $\downarrow$ & PSNR $\uparrow$ & SSIM $\uparrow$ & LPIPS $\downarrow$ & PVQA $\uparrow$ \\
    \hline
    \algoName & 0.027 & 63.865 & 0.997 & 0.096 & {\bf 72.91}\\
    \hline
    \algoName w/o planning & {\bf 0.024} & {\bf 64.362} & 0.997 & {\bf 0.092} & 71.74 \\
    \hline
    Planning w/ CSM~\cite{yu2021modular} & 0.028 & 63.777 & 0.997 & 0.093 & 72.53 \\
    \hline
    SVG-LP~\cite{denton2018stochastic} & 0.025 & 64.154 & 0.997 & 0.099 & 63.29  \\
    \hline
  \end{tabular}
  }
  \label{tab:baseline-ablation-metrics}
\end{table}

\subsection{Experiments}

For the experiments described, all models are performing keyframe prediction. Our hypothesis was that keyframe prediction would yield better, yet less smooth results than dense prediction as the prediction horizon is shorter. 

\vspace{3pt}
\noindent \textbf{Q1. \algoName without Cognitive Planning}. We trained our approach for 200 epochs for both dense as well as keyframe prediction. As a baseline video generation method lacking cognitive planning, we chose Stochastic Video Generation with a Learned Prior (SVG-LP)~\cite{denton2018stochastic}. A quantitative comparison between the two can be seen in \cref{tab:baseline-ablation-metrics}. 

We found that SVG-LP~\cite{denton2018stochastic}, lacking language direction, is struggling to even begin the task at times. The board in many cases remaining empty and in the keyframe prediction case, even larger objects in the scene, e.g. the robot arm, are blurry. Taking into account the possible combinations of letters and colors, as well the degree of stochasticity in actions and positions of the board and the letters, the complexity of the task seems to be prohibitive for SVG-LP~\cite{denton2018stochastic}. In comparison, \algoName is able to predict reasonable future frames that satisfy the language goals provided, despite the letters still lacking clarity.  

\begin{table}[h]
 \caption{OCR metric for ablation studies for final frame of keyframe prediction.}
\small
  \centering
  \resizebox{\linewidth}{!}{%
  \begin{tabular}{|c| |c| |c| |c|}
    \hline
    Architecture & Color (\%) & Letter (\%) & Letter \& color (\%) \\
    \hline
    \algoName & {\bf 51.67} & {\bf 6.735} & {\bf 4.095}\\
    \hline
    \algoName w/o planning & N/A & 3.179 & N/A \\
    \hline
    Planning w/ CSM~\cite{yu2021modular} & 22.737 & 3.017 & 1.078 \\
    \hline
    SVG-LP~\cite{denton2018stochastic} & N/A & 0.754 & N/A \\
    \hline
  \end{tabular}
  }
  \label{tab:OCR-metric}
\end{table}

\vspace{3pt}
\noindent \textbf{Q2. The value of GPT-2 Language Planner in \algoName.}
We assess the importance of a high level task planner and extensible language representations with two ablation studies. Firstly, we assumed absence of a planner and trained our video prediction approach using only high level task descriptions. That practically entails combining only the high level description with every visual observation during training, instead of a natural language instruction that corresponds to the action performed on that frame. Given our dataset design prevents trivial solutions in regards to inferring actions from the high level description of a task, we show that a modularized system architecture which contains a dedicated planner, performs better. Quantitative results can be seen in \cref{tab:baseline-ablation-metrics} where a qualitative example for the same task is demonstrated in \cref{fig:svg-See-PP-ablations}.

As expected, the presence of step-by-step goal directives in natural language greatly improves learning of the various concepts and actions. Relying purely on a general high-level task description for a task family like spelling provides little information about what is being asked. Assuming the high level task description suffices, we observe similar results to pure video predictions such as SVG-LP~\cite{denton2018stochastic} presented above, where the task cannot even begin. Interestingly, the task description alone seems to have equivalent assistive power to the learning process, as no language input at all.    

\begin{figure*}[thpb]
  \centering
  \includegraphics[width=\linewidth]{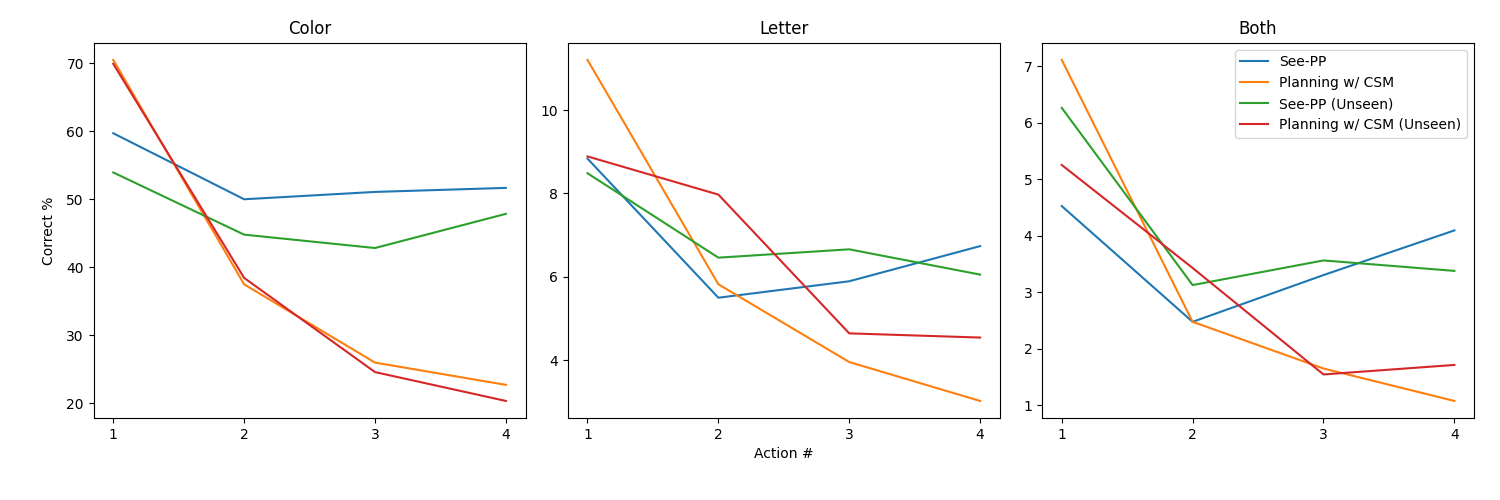}
  \vspace{-10pt}
  \caption{Keyframe-wise accuracies for \algoName, Planning w/ CSM~\cite{yu2021modular}, and corresponding seen-unseen experiments. The plot "Color" (left) shows the percentage of objects with the correct color placed in the correct location as described by the GPT-2 Language Planner, while "Letter" (center) shows the percentage of objects that are the correct letters. "Letter \& color" (right) represents the percentage of correct objects (both correct color and letter) according to the plans from the GPT Language Planner.}
  \label{fig:keyframe-fig}
\end{figure*}

Secondly, we assumed presence of a planner but opted for the original CSM~\cite{yu2021modular} structure which does not leverage extensible language representations. For this ablation study, we assumed a random train/test split of the dataset in order to fairly evaluate the impact of language embedding use towards concept grounding. As a reminder, a random train/test set ensures that, while individual videos are unique in the test set, all or most concepts that define the set of available concepts is known and shared between the training/test sets. This guarantees no missing concept knowledge for methods like the original CSM\cite{yu2021modular} that operate on pre-defined one-hot encodings. A qualitative comparison for the same task can be viewed in \cref{fig:svg-See-PP-ablations}. Our hypothesis was that for seen concepts at least, both representations should hold comparable information and thus perform rather similarly. However, we observed that our method still performs slightly better. We attribute this to the fact that language embeddings derived from pre-trained CLIP have been optimized to recognize visual concepts through appropriate association with language. 

\vspace{3pt}
\noindent \textbf{Q3. Generalization ability of \algoName}. 
In regards to the generalization, we once again employed our dataset split containing unseen concepts in the test set, as well as a random split version. For the purposes of this study, we omitted from the predefined concept dictionaries all letter and color concepts included in our unseen test split. Subsequently, we amended the original CSM\cite{yu2021modular} so that all unknown concepts get reduced to a 0 bit vector instead of introducing new encodings. A qualitative example of generalization to unseen concepts and how it compares can be seen in \cref{fig:seen-unseen}. As expected, our method is performing worse on unseen concepts compared to predicting on the random split where concepts may appear both on the training and test set. However, it can produce somewhat reasonable results compared to CSM's~\cite{yu2021modular} one-hot encoding representations which does not handle unseen concepts unless they are added into its pre-defined dictionary and the model is retrained. We hypothesize that reasonable actions can be attributed once again to CLIP embedding contrastive optimization between the visual and language modalities, which may already correctly be embedding particular concepts such as associating language with letter appearance. An interesting example we thought of was e.g. if images in CLIP’s training set included banners, billboards etc.

Finally, it is important to highlight that while we leveraged a purely language directed planning module~\cite{jansen2020visually} and Concept Slot Module as the action and concept grounding mechanism for video generation~\cite{yu2021modular}, different methods can be utilized. Our work focuses primarily on proposing a framework that can solve the cognitive planning task. 

While we believe that our framework can be leveraged to solve the cognitive planning problem, selecting an optimal planning and video generation module to achieve results is still an open problem. While simple, the planner we selected for this work is purely language-based and does not accept visual feedback. 
Such a purely language-based planner is static and does not perform replanning in the case of an inaccurate prediction. Experimenting with a closed-loop planning method that receives visual feedback from the current state in every step, and backtracks in failure cases is a direction towards addressing this issue. In addition, the video generation mechanism we have selected does not handle precision tasks such as spelling, and remains sensitive to egomotion and proportions of objects of interest in the scene compared to background. Lastly, this work was limited on a single simulation dataset with a certain number of properties, and was not evaluated on real world data which is our next step. Another exciting future direction involves use of such action grounded videos for low level robotic control through imitation learning approaches. 

\subsection{Semantic evaluation of method using OCR}

In order to capture the semantic utility of our models, we developed a customized OCR system which is comprised by two classifiers that identify the character and its color respectively. For the models which use planning in their pipeline, we evaluate three metrics for each action step: (i) color, percentage of objects whose color matches the one specified by the language planner; (ii) letter, the percentage of correctly placed letters, regardless of their colors; (iii) letter \& color, the percentage of correctly placed objects, in terms of both color and letter, as specified by the plan. The results for \algoName and Planning with CSM~\cite{yu2021modular}, as well as their variants that are trained on the seen split and tested on the unseen split, are presented in~\cref{fig:keyframe-fig}. For the models which do not support planning, we evaluate the percentage of letters correctly placed in the final frame, included in~\cref{tab:OCR-metric}. 

\subsection{Planner Evaluation}

As part of assessing the efficacy of our method, we performed an evaluation of the planner module separately. More specifically, we evaluated how well a pure language planner such as a fine-tuned GPT2 model used for this work, is able to produce valid plans for the spelling task of various words. The structure of the dataset used for fine-tuning GPT2 on the spelling domain has been described in the main paper while we also provide the training script used. Given that this type of planner does not receive visual feedback, a plan is considered valid if the series of language instructions predicted, indeed form the word that was instructed to be spelled by the high level task directive, e.g. ``\texttt{Spell the word PAIR}", irrespective of colors or whether the letter and color combination is present in the initial frame provided as input to our concept grounding video generator. For our test set containing 442 unique 4-letter words, we measured the action level success rate as well as the overall task level success rate of the planner. The action level success rate for our language input only planner was {\bf 93.72\%} while the overall task level success rate was {\bf 86.99\%}. This denotes that pre-trained language models with some fine-tuning on the domain, can be powerful planners, however the lack of visual feedback surely presents drawbacks that must be addressed in order to improve end-to-end performance of language-guided cognitive planning with video prediction which is also corroborated by recent work 
\cite{huang2022language}.

\section{CONCLUSION}
In this work, we focused on the problem of cognitive planning in video prediction and proposed a framework that performs task planning from high-level descriptions and subsequently, utilizes this plan towards “imagining” a sequence of future states that implement this plan. 
Our results highlight the importance of task planning for generating video sequences that accurately perform the task. In addition, we demonstrate that extensible language representations allow for better grounding generalization. With this work we aspire to showcase that cognitive planning is a promising direction not only for video prediction of actions that complete a task, but also for future trajectories that may be further leveraged for robotic low level control. 






\newpage
\bibliographystyle{IEEEtran}
\bibliography{see-pp.bib}

\end{document}